\documentclass[manuscript, screen]{acmart}
\pdfoutput=1 
\usepackage{booktabs} 
\usepackage{graphicx}  
\usepackage{subfig}
\usepackage{amsmath}
\usepackage{amssymb}
\usepackage{float}
\usepackage{xcolor}
\usepackage[normalem]{ulem}

\DeclareMathOperator*{\argmaxA}{arg\,max}
\usepackage[ruled]{algorithm2e} 

\setcopyright{acmcopyright}

\begin{document}

\title{Mitigating Class-Boundary Label Uncertainty to Reduce Both Model Bias and Variance}

\author{Matthew Almeida}
\affiliation{%
  \institution{University of Massachusetts Boston}
  \streetaddress{100 Morrissey Boulevard}
  \city{Boston}
  \state{Massachusetts}
  \postcode{02125}
}

\author{Yong Zhuang}
\affiliation{%
  \institution{University of Massachusetts Boston}
  \streetaddress{100 Morrissey Boulevard}
  \city{Boston}
  \state{Massachusetts}
  \postcode{02125}
}

\author{Wei Ding}
\affiliation{%
  \institution{University of Massachusetts Boston}
  \streetaddress{100 Morrissey Boulevard}
  \city{Boston}
  \state{Massachusetts}
  \postcode{02125}
}

\author{Scott Crouter}
\affiliation{%
    \institution{University of Tennesee}
    \city{Knoxville}
    \state{TN}
    \postcode{37996}
}

\author{Ping Chen}
\affiliation{%
  \institution{University of Massachusetts Boston}
  \streetaddress{100 Morrissey Boulevard}
  \city{Boston}
  \state{Massachusetts}
  \postcode{02125}
}

\renewcommand\shortauthors{Almeida, M. et al}

\begin{abstract}
 The study of model bias and variance with respect to decision boundaries is critically important in supervised classification. There is generally a tradeoff between the two, as fine-tuning of the decision boundary of a classification model to accommodate more boundary training samples (i.e., higher model complexity) may improve training accuracy (i.e., lower bias) but hurt generalization against unseen data (i.e., higher variance). By focusing on just classification boundary fine-tuning and model complexity, it is difficult to reduce both bias and variance. To overcome this dilemma, we take a different perspective and investigate a new approach to handle inaccuracy and uncertainty in the training data labels, which are inevitable in many applications where labels are conceptual and labeling is performed by human annotators. The process of classification can be undermined by uncertainty in the labels of the training data; extending a boundary to accommodate an inaccurately labeled point will increase both bias and variance. Our novel method can reduce both bias and variance by estimating the pointwise label uncertainty of the training set and accordingly adjusting the training sample weights such that those samples with high uncertainty are weighted down and those with low uncertainty are weighted up. In this way, uncertain samples have a smaller contribution to the objective function of the model's learning algorithm and exert less pull on the decision boundary. In a real-world physical activity recognition case study, the data presents many labeling challenges, and we show that this new approach improves model performance and reduces model variance.
\end{abstract}

%
%

\ccsdesc[500]{Computing methodologies~Knowledge representation and reasoning}
\ccsdesc[300]{Computing methodologies~Neural networks}

\keywords{Bias, Variance, Label Uncertainty, Neural Networks}

\maketitle

\section{Introduction}

One of the most important tasks in modern machine learning is that of supervised classification \cite{ng2017talk}, whereby a training set $\mathbf{X}$, with associated class labels $y$, is used to minimize the value of an objective function $L(\mathbf{X}, \theta)$ with respect to the data $\mathbf{X}$ and model parameters $\theta$, such that the trained model is able to reliably assign labels to new, unseen data.

Training a model that will generalize to unseen data is a fundamental challenge in supervised learning, and is subject to the bias-variance dilemma \cite{geman1992neural}\cite{shalev2014understanding}. To lower bias, the model needs to be adapted such that the decision boundary is permitted to contort to accommodate more boundary training samples and improve training accuracy, which results in a more complex classification model (see Figure 1(a)). However, in this process noisy or uncertain points may also be accommodated, which could harm the generalization and make predictions less accurate against a test set. On the other hand, a less complex, higher-bias model is relatively simple and may exhibit improved generalization (i.e., have a lower variance). One issue with reducing both model bias and variance lies in the trustworthiness of a sample: ideally, an informative sample should be wholly accommodated and a noisy sample should be discounted or discarded completely.

\begin{figure*}[t!]%
    \subfloat[Decision Boundaries in 2D classification]{\includegraphics[width=4.9cm]{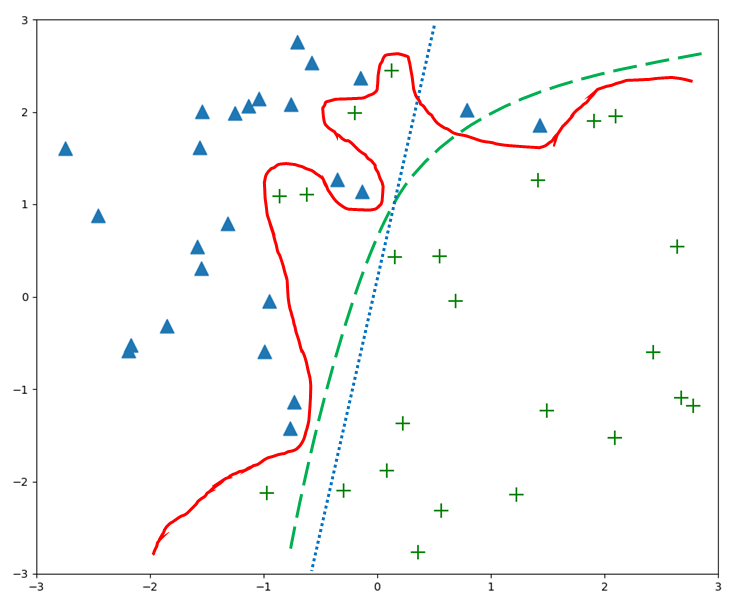}}%
    \subfloat[Generating Processes for MNIST digits]{\includegraphics[width=4.9cm]{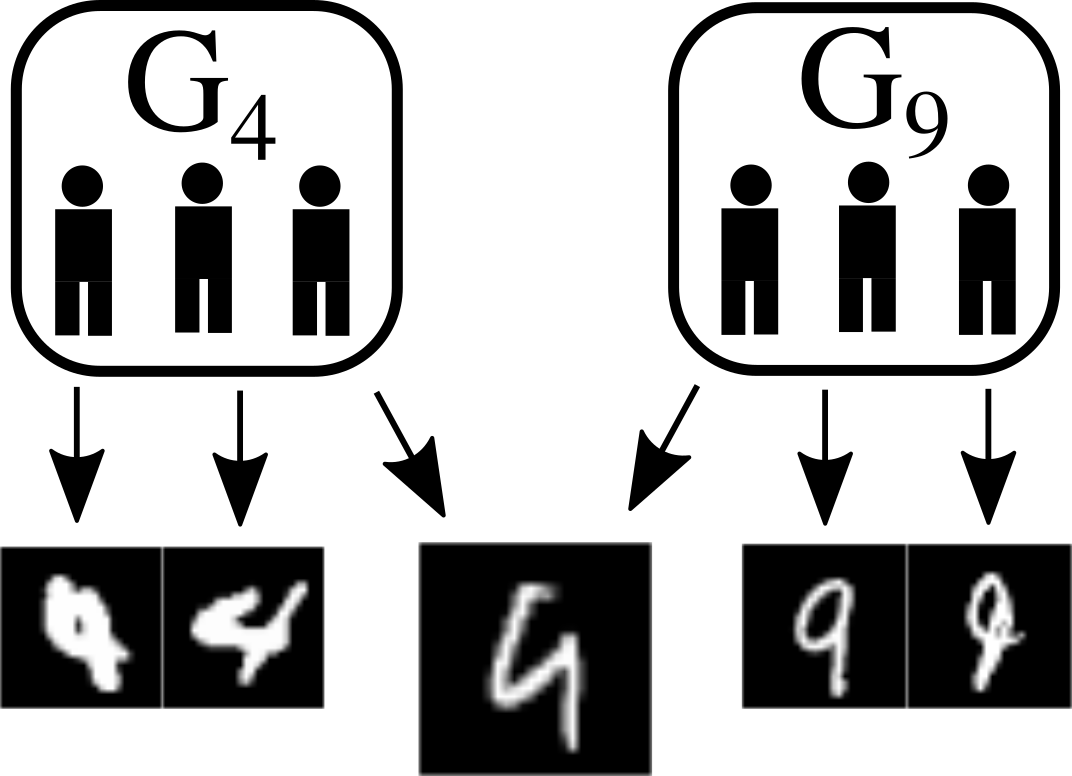}}%
    \subfloat[Example data from  activity recognition]{\includegraphics[width=4.9cm]{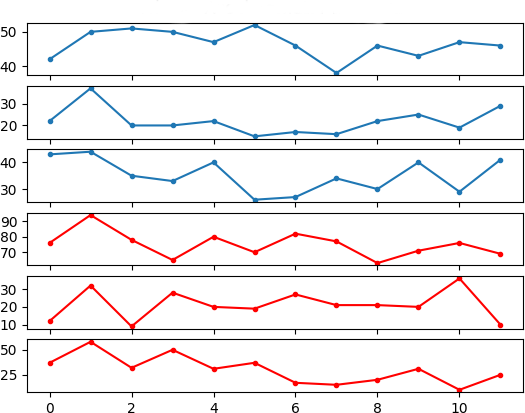}}%
    \caption{Three examples illustrating label uncertainty. (a) Three decision boundaries drawn by different classification models. The green (dashed) model is the true decision boundary, but it is impossible to know. The red (solid) model has adjusted to accommodate all samples and results in an overfitting low-bias but high-variance model. The blue (dotted) model is an underfitting low-variance but high-bias model. Uncertainty in the label of points near decision boundary could make a large difference when determining the boundary. (b) Represents the generating processes for "people writing 4s" $(G_4)$ and "people writing 9s" $(G_9)$. The central picture will be labeled either 4 or 9, but could be generated by either process. (c) shows two training examples from an activity recognition data set (collected on a hip-mounted tri-axial accelerometer). The top 3 time series (in blue) make up a walking sample, the bottom 3 (in red) a running. Such labeled samples are very difficult to differentiate.}
    \label{figure1}%
\end{figure*}

Understanding and study of uncertainty has long been an active topic in AI research. 
We take a data-centered view and consider that noise and uncertainty comes from four sources:

\begin{itemize}
\item{\textbf{Value noise}: value noise exists in collected values due to imprecise collection procedure and measurement tools, which is stochastic noise or aleatoric (inherent to the problem) uncertainty discussed in literature \cite{der2009aleatory}.}
\item{\textbf{Feature uncertainty}: failure to include sufficient discriminative features in the collecting process can result in overlapping classes that would be separable in the higher-dimensional space that would describe the data if the necessary features were included.}
\item{\textbf{Distribution uncertainty}: the number of samples is too small or data is biased, and so doesn't accurately reflect the true data distribution. This uncertainty is often called deterministic noise or epistemic (due to the modeling process)  uncertainty in literature \cite{der2009aleatory}.}
\item{\textbf{Label uncertainty}: class labels are often conceptual entities, and labeling is performed by human annotators. Even with carefully designed labels and experienced annotators uncertainty can arise due to different interpretation of labels and samples.}
\end{itemize}

This categorization of noise and uncertainty solely serves to inform our approach; a comprehensive discussion of noise and uncertainty goes beyond the scope of this paper. Here, we focus on a particular type of label uncertainty: that which is tied to the representation of the samples rather than errors which are stochastic in nature (labels that are flipped with a certain probability by a random process). As we will illustrate in the following sections, better understanding and handling of label uncertainty can contribute to the reduction of both model bias and variance. 

While label uncertainty can be revealed through multiple inconsistent labels if more than one annotator is used, in practice a sample is often annotated only once due to prohibitive labeling cost. Using a single label for both high- and low-confidence samples obscures the label uncertainty. In this paper we will consider this specific type of heteroscedastic (data-dependent) label uncertainty and provide a method to mitigate its effects on the decision boundary of the classification model. Our main idea is to use a $k$-nearest neighbor-based entropy measure to estimate the degree to which a point's label is uncertain: the label of a point surrounded by points of another class should be treated as less trustworthy - and the point's ability to move the decision boundary reduced - as the position of its representation is suspicious. Points in a highly heterogeneous neighborhood are similarly weighted down, as they are in an area of the representation space that is very chaotic; it is possible the points from two different classes are being projected down an unavailable data axis into the same region of the feature space.

To be more specific, our approach to improve both model performance and generalization is to identify points in the dataset that are likely to be "noisy" or "misleading" to the model (terms which will be defined formally in section 3 of this paper) due to labeling uncertainty and automatically adjust their corresponding sample weights such that data samples with uncertain labels do not contribute to the loss function to the same extent as informative data samples during training. We derive a sample weight for each point, using a function that calculates a pointwise score based on the entropy and distances of samples within each point's neighborhood, such that a point in a homogeneous neighborhood, with many neighbors of the same class as itself, will be weighted up, and points in heterogeneous neighborhoods will be weighted down as potentially noisy or mislabeled. The main contributions of this paper are the following:

\begin{itemize}
\item A novel definition of what is meant by an informative training sample with respect to its paired label. Instead of the existing approaches focusing on model fine-tuning - normally subject to the bias-variance tradeoff - we directly attack the core of the problem: how to measure the trustworthiness of a sample so we can decide whether a decision boundary should be adjusted to accommodate it accordingly. In this way we can  improve generalization performance while reducing model variance.
\item A new $k$-nearest neighbor-based framework for estimating label uncertainty point-by-point and mapping it to sample weights for use in model training.
\item When performing experiments, we configure our training environment such that our runs are deterministic for each random seed. In this way we can be sure that variation in performance is due to the variable we intend to observe, the sample weighting, not from randomness of learning models.
\end{itemize}

This paper is organized as follows. In the next section, we discuss related work. We then present our $k$-nearest neighbor-based label uncertainty measure and describe how to map those values to sample weights. We then describe our experimental procedure and results, and give final concluding remarks.

\section{Related Work}

\label{rel_work}
The tradeoff of bias and variance has been studied from different perspectives.  Geman, et al. introduced the dilemma in terms of neural networks (NNs) in \cite{geman1992neural}, showing that increasing variance with model capacity makes NN models require an "unrealistic" number of training examples that they could have not have foreseen becoming realistic with today's data collection and storage. Goodfellow et al. presented an updated discussion in \cite{Goodfellow-et-al-2016}. We examine the bias-variance relationship in settings where labeling errors are data-dependent, the probability of a given example $i$ having an incorrect label being dependent on the example's representation, $\mathbf{x_i}$. This setting could arise in human activity recognition, for example, when trying to classify walking and running: examples along a model's decision boundary separating the two classes are much more likely to be mislabeled than examples far from it (imagine a model that separated only on a person's average speed over a window of activity; examples near the speed representing the split point between the classes likely have much more uncertainty to their labels than very high- or low-speed examples). 

In \cite{liu2016classification}, Liu, et al. took an approach to handling label noise using importance-based sample reweighting. They worked specifically on 2-class label noise, where there is certain probability $\rho_{-1}$ that an example of class $-1$ will have its label flipped to a $+1$, and a probability $\rho_{+1}$ that an example of class $+1$ will have its label flipped to $-1$. They used a density ratio estimation method to perform the reweighting. This work and that of Scott, et al. \cite{scott2013classification}\cite{blanchard2010semi}\cite{scott2015rate} requires estimation of the noise rates $\rho_{-1}$ and $\rho_{+1}$. The authors of \cite{menon2015learning} offer optimization methods to avoid issues due to binary label noise and give steps to estimate noise from corrupted data. 

In \cite{northcutt2017learning}, the authors introduce a method called Rank Pruning to treat noise in labels, whereby they train one or more classifiers and then prune the training set of likely false positives and false negatives based upon the confidence rankings of the trained models; they then retrain a new model based upon the cleaned training set. With our method, we are concerned both with points that are actively misleading to the model (and should be removed entirely, as in the case of label corruption as in \cite{northcutt2017learning}), but also with classes with definitions that are more ambiguous, such as those in activity recognition; in these cases, removing points with low model confidence could result in a loss of useful information.

Natarajan, et al. in \cite{natarajan2013learning} also addressed binary classification with class-specific label noise, offering approaches to modify surrogate loss functions robust to it.

A popular avenue of current research studies a form of  stochastic label noise that is assumed to be as bad as possible for the model, when creating adversarial examples \cite{fawzi2016robustness} \cite{fawzi2018analysis}\cite{goodfellow2014explaining}\cite{gu2014towards}.

In \cite{ren2018learning}, the authors use sample weighting to improve the performance of deep learning models.  They develop a method to calculate sample weights for examples by learning a weighting during optimization, modifying the weights at each step based on how they reduce loss with respect to mini-batches drawn from a high-quality validation set. This method can help with label noise because instances with incorrect labels should presumably have very poor agreement with the validation set and therefore be weighted down. In our case, where we are trying to improve generalization along decision boundaries where we expect labeling problems, it would be difficult to obtain a truly clean validation set. Additionally, the authors analyze the method only on stochastic noise settings, where there is a uniform probability of label flipping or a certain probability with which labels from any class are flipped to a 'background' class, representing the case when human annotators miss a positive example. Other recent methods \cite{goldberger2016training}\cite{jiang2017mentornet} also assume stochastic label noise or corruption that is not data-dependent.

We use a $k$-nearest neighbor-based method to calculate our pointwise uncertainty scores, by comparing the local self-information of the class of each point $\mathbf{x_i}$ with the local self-information of the other classes in the dataset, weighted by the distances from each point to the other $k-1$ points.

$k$-nearest neighbor methods are commonly used in estimators of differential entropy and mutual information over continuous random variables \cite{berrett2016efficient} \cite{gao2018demystifying} \cite{gao2017density}. A popular method from \cite{kozachenko1987sample} uses the volume of open $d$-dimensional balls around each point, with a radius of the distance from the point to its $k$th neighbor, to estimate the pointwise local densities for the available samples, then uses those volumes to compute an estimate of the global entropy. 

We work in the discrete case, analyzing the local self-information over a finite set of classes, but find the $k$-nearest neighbor approach to entropy estimation valuable because it allows us to look at the local label entropy by class from a pointwise perspective. This gives us a measure of surprise to find the point's label at its location in the representation space, which we combine with the sparsity of the neighborhood and the local entropy of all classes to assign a score from which we can derive a sample weight.

In our work, we consider the multi-class case where there is uncertainty in the labeling. We will show how to reduce bias and variance together in the following sections.

\section{Overview of model bias and variance and connection with label uncertainty}
\label{biasvariance}
It is clear that label uncertainty impacts the trustworthiness of a sample, which in turn determines how much the sample should be accommodated when a decision boundary is produced as finding an optimal decision boundary can reduce both bias and variance. In this section we will discuss how we define label uncertainty, then analyze its connection to model bias and variance.

\subsection{Uncertain and informative data points}
\label{noisyinf}
In this paper, we make many references to the ideas of uncertain and informative data points. We use these terms relative to the ground truth of the classification problem and a hypothetical "ideal model" or true labeling function.  Any set of training data can be viewed as a set of draws from some unknown joint distribution, with each class representing a marginal distribution over the feature space. Each vector describing those samples is a single point in the feature space. 

We consider the case where the data collection and labeling process is complete and cannot be revisited. Of course, if more features could be added to each point's representation, the neighborhoods of the points would be changed and consequently the uncertainty estimation would be different as well. If adding a feature to a point maps it into a homogeneous region instead of a heterogeneous one, treating it as a more certain point is a reasonable approach.

A model is then a function $g(\mathbf{x}): \mathbb{R}^d \to \mathbb{N}$ that maps input vectors to class labels, integers that denote which class - which marginal distribution - a data vector was most likely generated from. It may be that a particular data point, say the MNIST \cite{lecun1998gradient} 9 with its top left open enough to resemble a 4 (as illustrated in Figure 1(b)), could have been generated by the process of people writing 9s with probability $p$ and the process of people writing 4s with probability $(1-p)$ (for the purposes of this discussion, we assume no one writing any of the other digits could ever produce the sample). When we have a label (in an ideal setting), we have the correct answer for which class marginal distribution a sample was generated from -- but we do not know whether or not that class's generating process was the one most likely to generate a sample at that point in the feature space.

In the cases where the same point in the feature space could be drawn from more than one class marginal distribution - there is some overlap - an ideal model, a model with perfect knowledge of the probability with which each process will generate a sample at each point in the space, cannot have perfect accuracy. The best that a model can do with such data is to predict the most likely generating class marginal distribution at each point in the domain. If the example digit is generated by the process producing 9s with probability .7 and the process producing 4s with probability .3, the best possible model can only predict 9 for that image, and be incorrect 30\% of the time.

This leads us to a formal definition of informative and uncertain data points: a data point is "informative" if it is of the class most likely to generate a data point at its location in the feature space. It is "uncertain" if it was generated by any other class. 

\textbf{Definition} Let $\mathbf{X}$ be a data set composed of $n$ $d$-dimensional training samples $\mathbf{x_i} \in \mathbf{X}$, and let $f(\mathbf{x_i}): \mathbb{R}^d \to \mathbb{R}^C$, where $C$ is the number of classes in the dataset, be a function taking any input point to the distribution over classes representing the probability that class $c$ generated sample $\mathbf{x_i}$. $\sum_{c} f(\mathbf{x_i})_c = 1$. Let $y \in \mathbb{R}^n$ give the observed labels for each $\mathbf{x_i}$. $f$ in this formulation can be seen as the ground-truth labeling function because it contains all possible knowledge of the labeling behavior of the problem.

Then we say that a sample $\mathbf{x_i}$ is an $informative$ point when $y(\mathbf{x_i}) = \textrm{max}[f(\mathbf{x_i})]$. In other cases, $\mathbf{x_i}$ is an $uncertain$ point.

\begin{figure*}[t!]%
    \subfloat[A case where downweighting examples results in lower generalization error]{{\includegraphics[width=4.9cm]{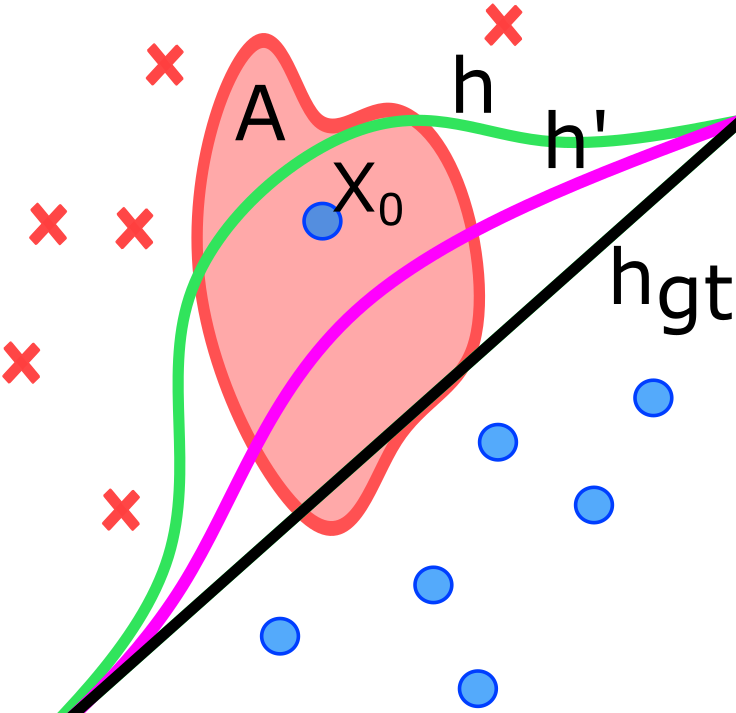} }}%
    \subfloat[A case where low model capacity causes fitting noisy points to be beneficial]{{\includegraphics[width=4.9cm]{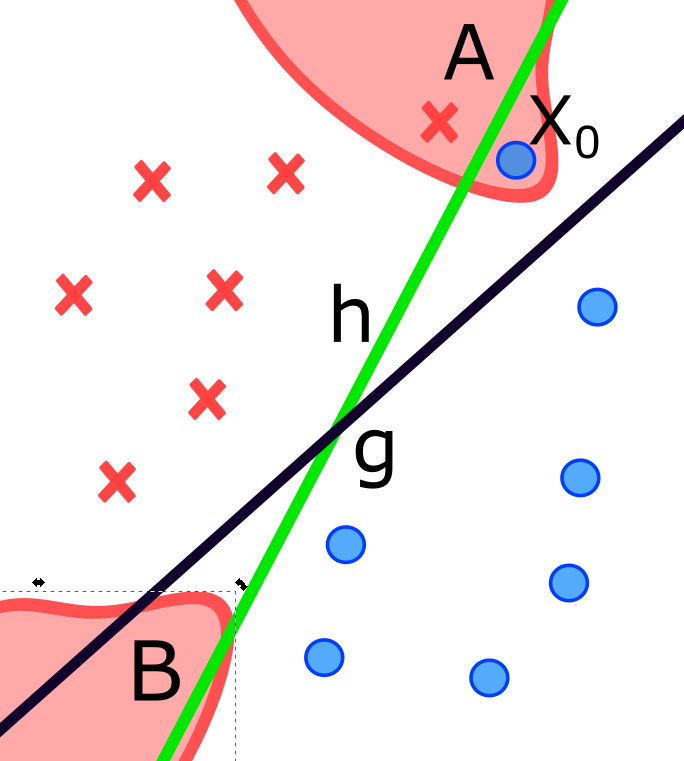} }}%
    \subfloat[A more detailed case where fitting noisy points helps in some areas and hurts others]{{\includegraphics[width=4.9cm]{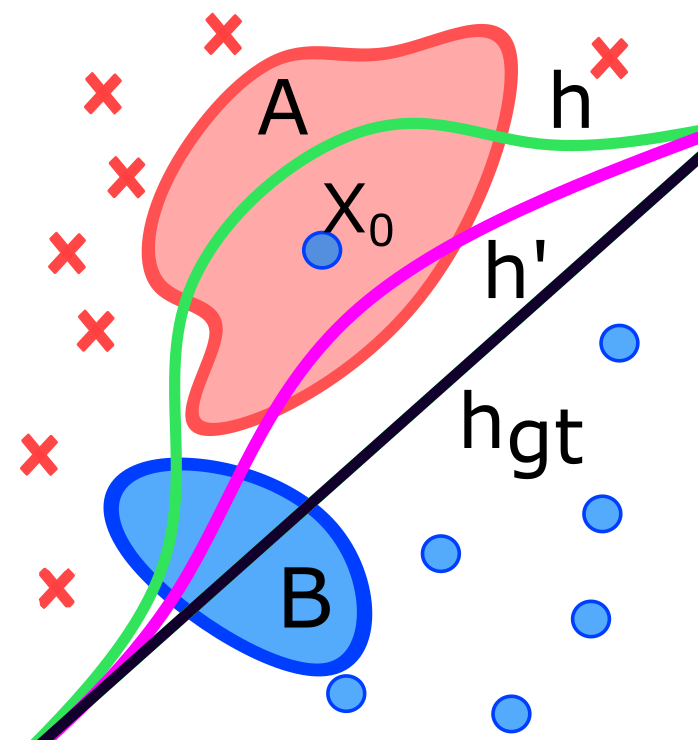} }}%
    \caption{Three examples illustrating behavior along decision boundaries. Each picture depicts the boundary between two classes, the red Xs and blue circles. The lines represent model decision boundaries and the shaded regions A and B represent areas that are colored for the color of the most likely class to be generated in that region. The red area is most likely to generate a red x, the blue to generate a blue circle. $X_0$ is a blue point in a red region, and is therefore a noisy point. (a) Three decision boundaries drawn by different classification models. $h$ accommodates noisy point $X_0$, and so any future points generated in the large region underneath $h$ will be classified incorrectly. $h'$ was trained with $X_0$ weighted down, so the erroneous region is smaller. $h_gt$ was trained with no noise, and is optimal (but imperfect). (b) Here, the downweighting process fails because of the low model capacity. $h$ accommodates a noisy point that model $g$ does not, but the mistake helps because all of region B is moved to the correct side. (c) A more complex case where accommodation of a noisy point causes misclassifications in A but correct classifications in B.}
    \label{figure2}%
\end{figure*}

\subsection{Model bias and variance}

When we refer to model bias and variance, we refer to the bias and variance terms in the decomposition of the expected out-of-sample (generalization) error of a classifier, as introduced in \cite{geman1992neural} and re-presented in many canonical texts including \cite{hastie2005elements} and \cite{Abu-Mostafa:2012:LD:2207825}. A good term-by-term explanation is available in \cite{vijayakumar2007tradeoff}.

Take a setting where we have a problem domain $\mathcal{D}$, which consists of a dataset, $\mathbf{X_\mathcal{D}}$ and associated labels given by a true labeling function $f(\mathbf{X}_\mathcal{D})$ which encapsulates an element of data-dependent label noise, as above: for a given point $\mathbf{x_i}\in\mathbf{X}_\mathcal{D}$, $f(\mathbf{x_i})$ is vector-valued, giving the probability distribution of observing a particular label $y_i \in \{1...C\} $ at $\mathbf{x_i}$. $|f(\mathbf{x_i})| = C, \sum_{c} f(\mathbf{x_i})_c = 1$.

We can think of the objects subscripted with $\mathcal{D}$ as being population-level; let $(\mathbf{X}, \mathbf{y}) \thicksim (\mathcal{D}; f)$ refer to drawing a particular $\mathbf{X} \subset \mathbf{X_\mathcal{D}}$ and $\mathbf{y} \thicksim f(\mathbf{X_\mathcal{D}})$ from $\mathcal{D}$. This yields an observed set of individual points $\mathbf{x_i}$ and associated labels $y_i$, with each $y_i$ drawn from the distribution $f(\mathbf{x_i})$. Model bias and variance is decomposed from expected generalization error taken over such draws. Let $g$ be a model function that yields predicted class $g(\mathbf{x_i}).$ The typical decomposition of the expected prediction error of a model trained on a single draw into bias and variance is expressed as follows, using the squared error loss function \cite{Abu-Mostafa:2012:LD:2207825}. $\mathbf{y}$ represents the labels $\{y_i\}$ drawn from the distributions $f(\mathbf{x_i})$:

\begin{equation}
\mathbb{E}_{pred}[g(\mathbf{X})] = \mathbb{E}_\mathbf{X}[(g(\mathbf{x})-\mathbf{y})^2]
\end{equation}
\noindent

and we write the expected error over potential observed datasets from domain $\mathcal{D}$ as

\begin{equation}
\begin{aligned}
\mathbb{E}_\mathcal{D}\left[E_{pred}[g(\mathbf{X})]\right] & = \mathbb{E}_\mathcal{D}\left[\mathbb{E}_\mathbf{X}[(g(\mathbf{X})-\mathbf{y})^2]\right] \label{eq2} \\
& = \mathbb{E}_\mathbf{x}\left[\mathbb{E}_\mathcal{D}[(g(\mathbf{X})-\mathbf{y})^2]\right]\\
& = \mathbb{E}_\mathbf{x}\left[\mathbb{E}_\mathcal{D}[g(\mathbf{X})^2] - 2\mathbb{E}_\mathcal{D}[g(\mathbf{X})]\mathbf{y} + y^2\right]
\end{aligned}
\end{equation}

\cite{Abu-Mostafa:2012:LD:2207825} observe that $\mathbb{E}_\mathcal{D}[g(\mathbf{x})]$ is an "average function" over trained models and denote it $\bar{g}$, then derive the model bias and variance:

\begin{equation}
\mathbb{E}_\mathcal{D}\left[E_{pred}(g)\right] 
= \mathbb{E}_\mathbf{X}\left[\mathbb{E}_\mathcal{D}[g(\mathbf{X})^2] - 2\bar{g}(\mathbf{X})\mathbf{y} + \mathbf{y}^2\right]
\end{equation}

\noindent
Adding in terms summing to 0, $-\bar{g}(\mathbf{X})^2 + \bar{g}(\mathbf{X})^2$,

\begin{equation}
\begin{aligned}
& \mathbb{E}_\mathcal{D}\left[E_{pred}(g)\right] \\
= & \mathbb{E}_\mathbf{x}[\underbrace{\mathbb{E}_\mathcal{D}[g(\mathbf{X})^2] -\bar{g}(\mathbf{X})^2}_{\mathbb{E}_\mathcal{D}[(g(\mathbf{X})-\bar{g}(\mathbf{X})^2]} + \underbrace{\bar{g}(\mathbf{X})^2 - 2\bar{g}(\mathbf{X})\mathbf{y} + \mathbf{y}^2}_{(\bar{g}(\mathbf{X}) - \mathbf{y})^2}]\label{eq4}
\end{aligned}
\end{equation}

\noindent
where $(\bar{g}(\mathbf{X}) - \mathbf{y})^2$ is the bias and $\mathbb{E}_\mathcal{D}[(g(\mathbf{X})-\bar{g}(\mathbf{X}))^2]$ is the variance. Extension to loss functions beyond squared error and detailed analysis of systemic and variance effects is available in \cite{james2003variance}

\subsection{Bias, variance, and label noise}

The above shows a bias-variance decomposition for squared error in a setting where $\mathbf{y}$ is considered the absolute truth, not a particular draw from a set of data-dependent label distributions $f(\mathbf{X})$. Here we examine how label noise impacts the bias and variance of models.

Take a model function $h(\mathbf{X})$ trained on data with noisy labels, $(\mathbf{X}, \mathbf{y}) \thicksim (\mathcal{D}; f)$. Let $y_{gt}$ be the ground-truth label vector, the draw from $f(\mathbf{X})$ for which each point is assigned its most probable label: $y_{gt_{i}} = \argmaxA_c f(\mathbf{x_i})$, and let $h_{gt}(\mathbf{X})$ be a model trained on $y_{gt}$ of the same hypothesis class as $h$.

We can split the observed data $\mathbf{X}$ into two subsets, the informative points $\mathbf{X}_{info}$ and noisy points $\mathbf{X}_{noisy}$. $\mathbf{X}_{info}$ and $\mathbf{X}_{noisy}$ are disjoint and their intersection includes all examples in $\mathbf{X}$. Assume $\mathbf{X}_{noisy} \neq \emptyset$. In the case where we have a deterministic training process and infinite model capacity in the hypothesis class, $h_{gt}(\mathbf{X})$ will be Bayes optimal but $h(\mathbf{X})$ will be sub-optimal - it will incorrectly predict each point in $X_{noisy}$, and its bias relative to $h_{gt}$ will be higher.

If we were able to identify which $\mathbf{x_i}$ were in $\mathbf{X}_{noisy}$ via some process with perfect confidence, we could remove that subset from the training set and remove the effect of the label noise. Since we assume that any such attempt to identify noisy would have its own uncertainty, or that the labels themselves might not be perfectly orthogonal, we instead weight down those examples that we suspect are noisy. A model $h'(\mathbf{X})$ trained with the noisy $\mathbf{y}$ vector but with the noisy points downweighted will have a decision boundary between that of $h$ and that of $h_{gt}$. In this case, since potential $h'$ decision boundaries do not reach to the noisy points to the same extent the $h$ models do, the $h'$ models will have lower variance than the $h$ models - the models with the noisy points weighted down have better bias and variance.

Of course, the above does not hold in general, even if we stipulate that the points in $\mathbf{X}_{noisy}$ can be reliably identified. We assume above that we have deterministic training and infinite capacity - it is easy to imagine a case where a linear decision boundary is pivoted to accomodate a noisy point and the boundary on the far side of the pivot changes the prediction associated with a new area of the feature space from incorrect to correct; the incorrect accomodation of the point in this case would improve the model's performance. Additionally, improved variance does not guarantee better expected generalization error in all cases \cite{james2003variance}; a higher-variance model can have a lower variance effect on expected prediction error than a lower-variance model. Mislabeled points could also drag the decision boundary over regions that were previously being predicted incorrectly (possibly even because no data had been observed there), improving the expected prediction error over the whole domain.

With that said, many modern machine learning methods - especially neural networks - have enormous model capacity; two-layer neural networks can approximate arbitrary functions in the infinite-width limit \cite{leshno1993multilayer}. We expect that when working with data from real-world distributions, when we weight down points with high label uncertainty we will obtain models with improved decision boundaries in practice.

\section{Estimating pointwise label uncertainty}

\subsection{Requirements for the uncertainty estimation function}

Of course, to say for certain which points are uncertain and informative using the above definition (\S\ref{noisyinf}) would require knowledge of the generating processes or other information that could be difficult or impossible to obtain. Instead, we estimate which points are likely to be uncertain by examining each sample's neighborhood within the available dataset, defined by a parameter $k \in \mathbb{N}$, the neighborhood size (by number of neighbors, including the point itself).  We define a scoring function to assign a value to each point based on the entropy of observed classes within its neighborhood and the relative sparsity of the neighborhood, with the intention that the value is indicative of the uncertainty of the point's label. The score should have the following properties:

\begin{enumerate}
\item A sample should have score 0 when all $k-1$ neighbors are of the same class as the sample.
\item Examples in highly heterogeneous neighborhoods (i.e., neighborhoods with a high number of classes present) should have higher scores than points in homogeneous neighborhoods consisting of mostly their own class, but lower scores than points in homogeneous neighborhoods consisting of points of mostly another class.
\item Examples in relatively dense neighborhoods should have higher scores than points in relatively sparse neighborhoods, with label composition held constant.
\end{enumerate}

The intuition for these requirements follows from our goal to use only the information contained in the dataset, i.e., the neighborhood of each sample, to estimate its label uncertainty. If all other points in a given point's neighborhood are of the same class as the point, we choose to trust its label. Its neighborhood score should be 0, indicating no uncertainty; If a point is in a dense region of the feature space and its neighbors are all of another class, we should be highly suspicious of its label \sout{as} being potentially incorrect. The second and third requirement follow from how we think of noisy regions. Points with highly diverse labels in their neighborhood - especially in a dense neighborhood - are more likely to not be of the class most likely to generate a sample at that point in the domain, because the presence of many classes in the same neighborhood indicates that several class processes could generate samples in that region and, consequently, the model should put less weight on such samples when drawing the decision boundary. Performance gain from adjusting to accommodate those points is unlikely to generalize because the region is chaotic. Both informative and uncertain samples near class boundaries will have nonzero scores, as they will have neighbors with different labels.

\subsection{Incorporating Neighborhood Uncertainty Scores into the Loss Function for Classification}

After a neighborhood is analyzed from the view of label uncertainty, we need take a further step to perform classification. Uncertainty scores are converted to sample weights via a logistic mapping function and incorporated into the objective function optimized during model training: Let $\mathcal{L}(\mathbf{X},\Theta)$ denote the objective function without sample weighting, where $\mathbf{X}$ is the set of all data points $(\mathbf{x_i}, y_i), i \in [0,N)$ and $\Theta$ represents the model parameters. If $\mathbf{b}$ is the length-$n$ vector containing the neighborhood scores for each $(\mathbf{x_i}, y_i) \in \mathbf{X}$, and $g(\cdot)$ is the logistic mapping function taking neighborhood scores to sample weights, then our objective function becomes:

\begin{equation}
\mathcal{L}^*(\mathbf{X}, \Theta) = \frac{1}{N}\sum_{i=1}^{N}g(b_i)\mathcal{L}(\mathbf{x_i}, \Theta)
\end{equation}

\subsection{Calculation of Neighborhood Uncertainty Scores}
\label{nsformula}

We calculate the score for a sample $\mathbf{x_i}$ with label $y_i$ as follows:

\begin{equation}
\label{scoreequation}
b_{\mathbf{x_i}} = \frac{-C*(\frac{k_{y_i}}{k}\log\frac{k_{y_i}}{k} * \frac{k_{y_i}}{\sum\mathbf{d_{x_i}}})}{-\sum\limits_{j=1}^C(\frac{k_j}{k}\log\frac{k_j}{k} * \frac{k_j}{\sum\mathbf{d_j}})}
\end{equation}

\noindent
where $C$ is, as before, the number of classes in the dataset, $k$ is the number of neighbors that we consider for each sample, $k_{y_i}$ is the number of neighbors with the same label as $\mathbf{x_i}$, and $k_j$ is the number of neighbors with class label $y_j$. $\mathbf{d_{x_i}}$ is a distance vector that stores the normalized distances to the $k_{y_i}$ neighbors with the same class label as $\mathbf{x_i}$. The normalization is performed by setting the distance to $\mathbf{x_i}$'s nearest neighbor to be 1, and scaling the distances to the other neighbors based on that value. The terms $\frac{k_{y_i}}{\sum\mathbf{d_{x_i}}}$ in the numerator and $\frac{k_j}{\sum\mathbf{d_j}}$ in the denominator are included to weight the class self-information in the formula by the average inverse distance to the neighbors of that class (to reduce the influence of far-away points). The denominator of this formula equals 0 if all neighbors of the sample $\mathbf{x_i}$ have label $y_i$. In these cases, we define the value of $b_{x_i}$ to equal $0$. 

For a point $\mathbf{x_i}$, this calculation compares the entropy of labels in $\mathbf{x_i}$'s class,  $y_i$, to the expected entropy of labels in the neighborhood. This meets our established requirements:

\begin{enumerate}
\item When all $k-1$ neighbors have the same label, the entropy in the denominator of equation (1) is 0, and $b_{x_i}$ is defined to be $0$.
\item In a neighborhood that is highly heterogeneous, the total number of points with each label is similar (if one class label had many more points than the others, the neighborhood would not be highly heterogeneous). Therefore, the entropy term for the label of point $\mathbf{x_i}$, $\frac{k_{y_i}}{k}\log\frac{k_{y_i}}{k}$, is close to the expected entropy over all labels, represented in the denominator of equation (1). Additionally, inverse average distances are similar over all classes in such regions, so the neighborhood score is close to 1 for each point in the neighborhood, giving us the desired effect.
\item By weighting the terms corresponding to each class $j$ by the inverse average distance from $\mathbf{x_i}$ to its neighbors in class $j$, we reduce the effect of sparsely represented classes in the neighborhood and increase that of denser classes.
\end{enumerate}

\begin{figure*}
    \centering
    \subfloat{\fbox{{\includegraphics[width=4.7cm]{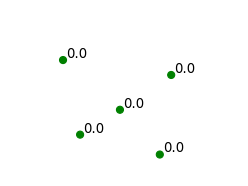} }}}%
    \subfloat{\fbox{{\includegraphics[width=4.7cm]{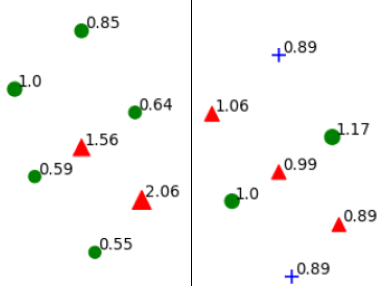} }}}%
    \subfloat{\fbox{{\includegraphics[width=4.7cm]{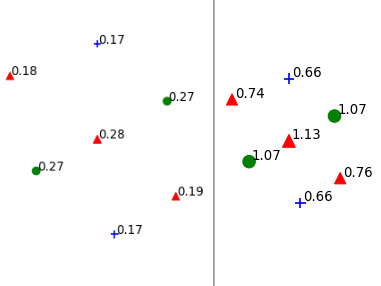} }}}%
    \caption{Three example figures demonstrating neighborhood scores, calculated in the 2D plane with Euclidean distance. Each corresponds to one of the requirements we have of a scoring function. The far left figure shows a neighborhood where each point has the same label; all are assigned a score of 0.  The central figure depicts two distinct neighborhoods with identical spacing but different label distributions. The neighborhood with a more even mix of labels represented has a more even distribution of scores. The right figure shows two copies of the same neighborhood, with the  relative spacing held constant but the distances between points increased by a constant factor. (Spacing increased 4x for score calculation, pictured at 2x for readability.)}%
    \label{fig:example}%
\end{figure*}

\noindent
Example values can be found in Figure (\ref{fig:example}).

\subsection{Mapping neighborhood scores to sample weights for classification}
\label{mappingscores}
There are potentially many ways to map neighborhood scores to sample weights.  The neighborhood scoring function has a minimum value of 0 (for a point in a fully homogeneous neighborhood).  

The logistic function (and especially the sigmoid function, a special case of the logistic) is commonly used in machine learning as a 'squashing' function to ensure output values fall in a certain range \cite{Gershenfeld:1999:NMM:299882}; we use a negative logistic function to transfer neighborhood scores into the desired range for sample weights, because it allows us to generate high weights ($>$ 1.0) on points with a low neighborhood score and low weights ( $<$ 1.0) on points with score near $\log_2 k$. This is exactly the behavior we want - points with low uncertainty are weighted up, and high uncertainty are weighted down. See the experiments section for further examination of this relationship. 

The hyperparameters that define this function do need to be tuned based on data. We find that logistic functions of form similar to the following fit our requirements:

\begin{equation}
\label{weightequation}
g(b_i) = \frac{\gamma}{1 + e^{-\alpha(-b_{x_i} + \beta)}} + \eta
\end{equation}

Here, $\beta$ controls the value of neighborhood score that the logistic function is centered on. We find empirically that the median of the nonzero scores is a good initial value for this parameter, and tends not to needlessly down-weight useful samples by considering too many of them to be uncertain. $\alpha$ controls the steepness of the logistic curve, the "hardness" of the threshold that separates an informative point that is up-weighted from a uncertain point that is down-weighted. $\gamma$ and $\eta$ take the score values and map them to values in the range $[\eta, \eta + \gamma]$, such that the low values of scores (near $0$) are mapped to nearly $(\eta + \gamma)$, and high values are mapped to $\eta$. 

We find empirically that the sample weights should fall in a range from $\approx 0.25$ to $\approx 2.0$; allowing weights to go to 0 effectively shrinks the available data set, reducing performance. Upweighting samples beyond 2.0 tends to overfit those samples too much when applied to our dataset.

\begin{table*}
\centering
\caption{Results of grid search}
\label{gridsearchresult}
\scriptsize
\begin{tabular}{@{}ccclccc@{}}
\toprule
\multicolumn{7}{c}{Top 5 Weight Combinations by Average Improvement Over Baseline (\%)}                                                                                                                        \\ \midrule
Group Assignments    & G0 / G1 / G2      & \multicolumn{1}{l}{Avg Over Baseline} &  & Group Assignments                     & G0 / G1 / G2      & \multicolumn{1}{l}{Avg Over Baseline} \\ \cmidrule(r){1-3} \cmidrule(l){5-7} 
NB Score   & 1.5 / 0.6 / 0.25  & +0.92                                  &  & Random                                & 2.0 / 0.6 / 1.0   & +0.74                                  \\
NB Score   & 2.0 / 0.6 / 0.25  & +0.91                                 &  & Random                                & 1.5 / 0.25 / 0.25 & +0.71                                  \\
NB Score   & 0.25 / 1.0 / 0.6  & +0.85                                  &  & Random                                & 0.6 / 0.25 / 0.25 & +0.69                                  \\
NB Score   & 0.25 / 0.6 / 0.25 & +0.84                                  &  & Random                                & 1.5 / 0.25 / 0.6  & +0.68                                  \\
NB Score   & 0.6 / 1.5 / 0.25  & +0.77                                  &  & Random                                & 1.5 / 0.25 / 2.0  & +0.68                                  \\
\multicolumn{1}{l}{} &                   & \multicolumn{1}{l}{}                    &  & \multicolumn{1}{l}{}                  &                   & \multicolumn{1}{l}{}                    \\ \midrule
 \multicolumn{7}{c}{Bottom 5 Weight Combinations by Average Performance Reduction from Baseline (\%)}                                                                                                                     \\ \midrule
Group Assignments    & G0 / G1 / G2      & \multicolumn{1}{l}{Avg Under Baseline} &  & \multicolumn{1}{l}{Group Assignments} & G0 / G1 / G2      & \multicolumn{1}{l}{Avg Under Baseline} \\ \cmidrule(r){1-3} \cmidrule(l){5-7} 
NB Score   & 0.6 / 0.6 / 2.0   & -3.7                             &  & Random                                & 0.6 / 0.6 / 0.6   & -0.82                                  \\
NB Score   & 0.25 / 2.0 / 2.0  & -3.7                                  &  & Random                                & 0.25 / 2.0 / 0.6  & -0.86                                  \\
NB Score   & 0.25 / 0.6 / 2.0  & -4.5                                  &  & Random                                & 1.5 / 2.0 / 1.5   & -0.87                                  \\
NB Score   & 0.25 / 0.25 / 1.5 & -5.4                                  &  & Random                                & 0.25 / 1.5 / 1.5  & -1.1                                 \\
NB Score   & 0.25 / 0.25 / 2.0 & -8.6                                  &  & Random                                & 0.6 / 0.25 / 1.5  & -1.9 \\ \toprule                                
\end{tabular}
\scriptsize
\end{table*}

\section{Experiments}
\subsection{Case Study 1: Classifications of A Real-World Physical Activity Dataset}
Objective and Accurate measurement of physical activity is a critical requirement for a better understanding of the relationship between sedentary behaviors, physical activity and health \cite{crouter2015estimating}\cite{mu2014bipart}. We evaluate our method on a physical activity recognition dataset collected from hip-mounted, tri-axial accelerometers from a cohort of 184 child participants. There were 98 male subjects from ages 8 to 15, and 86 female subjects from ages 8 to 14.  Each subject was observed for a period of lying rest with median 17 minutes (maximum 30 minutes), and median 4 minutes for each other activity (maximum 10 minutes). Researchers observed each activity and recorded the activity performed and the start and end times of each bout, so the data features ground-truth segmentation. We split each bout into discrete 12-second windows of activity described with the output of a single tri-axial accelerometer running at 1 Hz, resulting in 36 features per sample. We have 11,543 samples, and calculate the neighborhood scores using $k=5$ and cosine similarity.

Like many real-world applications, the labeling process is difficult and comes with significant uncertainty. Labeling is performed based on both in-person and video observation, and classes are often difficult to distinguish. There are 5 classes in our analysis: sedentary, light household and games, moderate-vigorous household and sports, walking, and running, and they are superclasses of the full label set, which consists of Computer Games, Reading, Light Cleaning, Sweeping, Brisk Track Walking, Slow Track Walking, Track Running, Walking Course, Playing Catch, Wall Ball, and Workout Video. Even among the superclasses, there are typically samples from different classes that appear to be very similar (e.g., walking across the house during a "light household" sample and walking across a basketball court during a "sports" sample). Using the even more fine-grained labeling approach would introduce more noise and drastically reduce the amount of data available per class, making it impractical.

\subsubsection{Grid search validation for neighborhood scoring function}

In our first set of experiments we aim to validate our scoring function, and show that the samples with high scores are in fact the uncertain samples and that weighting them down improves performance. To do this, we calculated the neighborhood score $b_{x_i}$ for each sample in the training set, and assigned those samples into groups. First, we put all samples $x_i$ for which $b_{x_i}= 0$ into group $G0$; a score of 0 indicates that a point's entire neighborhood is from the same class as itself. This is the zero-uncertainty group: our method considers points in fully homogeneous regions to have no label uncertainty. We then take all the remaining points, sort them by score, and divide them in half such that we have two more groups, $G1$, those points with low-but-nonzero scores, and $G2$, those points with the highest scores. Points that are close to decision boundaries but are not uncertain (by the definition in \S \ref{noisyinf}) should have low, nonzero scores, and points that are misleading should have high scores, so we aim to separate points around class boundaries into informative and uncertain points with this split. We perform this process to test if there is an advantage to downweighting, leaving the same, and upweighting the three groups split by estimated uncertainty; we need to make sure that our intuition to up-weight samples with low uncertainty and down-weight samples with high uncertainty holds in practice. We refer to experiments performed on data split this way as neighborhood or NB-weighted experiments. Of course, no test examples are ever used in the calculation of the uncertainty scores; inclusion of those examples would leak information about the test set into the training process whenever a test example was in the neighborhood of a training example.

\begingroup
\renewcommand{\arraystretch}{1.25}

\begin{figure*}[t]
  \begin{minipage}[c]{0.7\textwidth}
    \scriptsize
    \includegraphics[width=\textwidth]{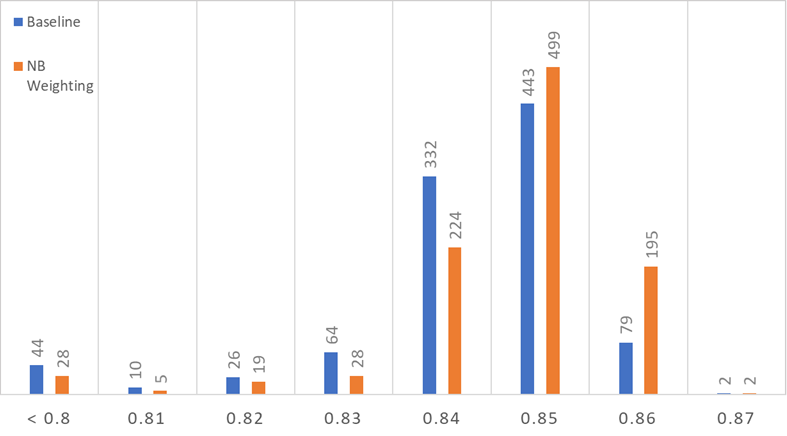}
    \scriptsize
  \end{minipage}
  \begin{minipage}[c]{0.18\textwidth}
        \label{subtable}
        \footnotesize
        \begin{tabular}{ccc}
        Accuracy & Baseline & NB Weighted\\
         (\%) & Models & Models \\
        $\leq$ 80 & 44 & 28 \\
        81 & 10 & 5 \\
        82 & 26 & 19 \\
        83 & 64  & 28 \\
        84 & 332 & 224 \\
        85 & 443 & 499 \\
        86 & 79 & 195 \\
        87 & 2 & 2 
        \end{tabular}
        \footnotesize
  \end{minipage}
  \caption{\textbf{Histogram of $k$-nearest-neighbor weighting vs. baseline}. After validating the approach with the grid search, we take the best sample weighting combination (1.5 / 0.6 / 0.25) and use equation (2) to create a continuous mapping of scores to weights, such that the continuous mapping follows the same weight pattern (low scores mapped to weight 1.5, high to 0.25). With $\gamma=1.25$, $\alpha=4$, $\beta=1.13$, and $\eta=0.25$, we run 1,000 pairs of models, each pair being one with neighborhood weighting and one with all weights set to 1.0, and both using the same random seed - model pairs see the same examples in the same order throughout training, and start from identical weight initializations. The models trained with $k$-nearest-neighbor weighting are better (clustered further right) on average, and exhibit lower variance over 1000 runs (baseline models have $\sigma^2 = 4.71 $ and $\sigma = 2.17\%$. NB-Weighted models have ($\sigma^2 = 2.53 $ and $\sigma = 1.59\%$)}
  \label{perfbytype}
\end{figure*}
\endgroup

For comparison we create a second split into three groups (such that the sizes of the groups correspond to the sizes of the NB-split groups) but assign samples to groups randomly. We refer to experiments with these splits as having been run with "random assignment" groups. We do this to make sure that the results we observe are due to our method and not to chance. By running the whole suite of experiments a second time on randomly split groups, we can observe the distribution of results from the random assignments to see what variations in performance we should expect due to randomness. We can then compare our NB-weighted results to make sure they are significant.

We perform a grid search to evaluate each possible combination of sample weight assignment and groups, using five discrete weights chosen to cover a range of weighting options but not leave large gaps: 0.25, 0.6, 1.0, 1.5, and 2.0. These values are chosen as proxies for the following possible ways to adjust the sample weights for a group: strongly downweight; somewhat downweight; no adjustment; somewhat upweight; strongly upweight.

This experiment is intended to show two main points: 1) that the score values capture useful information about the data's feature space and 2) that our interpretation of the score values is consistent with observed performance differences in the grid search; i.e., that weighting up the zero-score group (those points we identify as having no label uncertainty) and weighting down the high-score group (those points we identify as having uncertain labels) outperformed other sample weight-group assignments. The five weights were assigned to the three sample groups, for both assignment schemes, in all possible combinations. Each combination was run 10 times and the results (as measured against a fixed held-out test set) were averaged, for a total of 2,500 model runs ($5^3=125$ total combinations of weight assignments, $*10$ runs per assignment, $*2$ experimental conditions per combination-assignment=2,500 runs).

Training was performed using the Keras library \cite{chollet2015keras} and the Theano backend using single-threaded CPU computation only. This step was taken to remove nondeterminism introduced by multi-threaded CPU context switching and CuDNN. With these settings, a run with fixed sample weights and random seed is deterministic, and will finish training with the exact same result each time. 10 random seeds were generated once at the beginning of the experiment and the same 10 seeds were used for each combination of sample weight assignments to reduce the effect of particular combinations having stronger performance due to a lucky set of initializations within the weight space. We use a simple 2-layer Multi-Layer Perceptron architecture to keep running time reasonable.

The results of this process are shown in Table \ref{gridsearchresult}.  Reported figures represent change in model performance when using various weighting configurations compared to a baseline model that was trained with no sample weighting (all weights = 1.0).  The performance of the baseline model was 83.4\%, averaged over 10 runs. While there is no discernible pattern in the random results, as we would expect, there is a clear pattern in the results when $k$-nearest neighbor-based weighting is used: performance is strong when the most uncertain points (group 2) are weighted down (G2 is weighted down in all 5 of the top combinations), and performance is weak when the uncertain points are weighted up (G2 weighted up for all 5 of the bottom combinations). All 5 of the best neighborhood score combinations are better than the best one when weights are randomly assigned; all 5 of the worst weight combinations perform worse than the worst run under the random setting. This validates our interpretation of the neighborhood scoring function - weighting down the points with highest scores improves model performance over baseline, weighting up high-scoring points increases model focus on points with uncertain labels and decreases accuracy.

\subsubsection{Evaluating $k$-nearest neighbor weighting against baseline for activity recognition}

Our second round of experiments takes the best k-nearest-neighbor weighting model with weights (1.5 / 0.6 / 0.25), and uses equation (3) to create a continuous mapping function from scores to weights, then measures the performance of models trained with these weights against baseline models (where all sample weights=1.0) more thoroughly. We choose 1,000 random seeds from integers in the interval [0, 100,000) and run an NB-weighted model and a baseline model for each one, under the same CPU-based calculation conditions as the models from the grid search, so that any difference in performance is directly attributable to the difference in the weighting scheme. This yields 1,000 NB-weighted models and 1,000 baseline models. We summarize the results in Figure \ref{perfbytype}. The histogram on the left of Figure \ref{perfbytype} shows a distribution of trained models by performance, with better-performing models on the right. The absolute counts are provided in the table to the right of the histogram. We can see that the models using k-nearest neighbor weighting have both improved performance on average (are further right) and have lower variance (are more clustered in the histogram). The NB-weighted models are on average +0.534\% better than the baseline models by accuracy, and have greatly reduced variance $\sigma^2 = 2.53 $ and standard deviation $\sigma = 1.59\%$, as compared to the baseline models' variance $\sigma^2 = 4.71 $ and standard deviation $\sigma = 2.17\%$. Note that this is the variance of the model results, and is a different mathematical quantity than the model variance defined in \S\ref{biasvariance}; that variance is measured over models each trained on a different dataset sampled from a particular data domain. The variance calculated here is a variance over the model training process using a single dataset.

\section{Conclusion and future work}

With an eye on the bias-variance dilemma, we formulated a $k$-nearest neighbor-based method to estimate pointwise uncertainty in labeling and mitigate its effects by weighting down the samples in areas of the feature space with high density and label entropy. By working on the fundamental issue of the bias-variance dilemma (i.e., whether a decision boundary should accommodate a sample point according to the trustworthiness of that sample), we show improved model bias and variance in a real-world application. Using a neural network architecture to classify accelerometer data for activity recognition, we improve performance in a real-world domain where accurate, consistent labeling is very difficult. In future work, we hope to improve the method we use for estimation of label uncertainty, with the aim of obviating the need to calculate a distance matrix with $k$-nearest neighbors.

\begin{acks}
Funding for this research was provided by NIH grant 1R01HD083431-01A1: Novel Approaches for Predicting Unstructured Short Periods of Physical Activities in Youth. This research was also partially supported by the Oracle grant to the College of Science and Mathematics at the University of Massachusetts Boston.

\end{acks}

\bibliographystyle{ACM-Reference-Format.bst}
\bibliography{refs.bib}

\end{document}